\documentclass[10pt,twocolumn,letterpaper]{article}

\usepackage{graphicx}
\usepackage{booktabs} 
\usepackage{amsmath}
\usepackage{amssymb}
\usepackage{multirow}
\usepackage[table]{xcolor}
\usepackage{makecell}

\usepackage[pagenumbers]{cvpr} 

\definecolor{cvprblue}{rgb}{0.21,0.49,0.74}
\usepackage[pagebackref,breaklinks,colorlinks,allcolors=cvprblue]{hyperref}

\newtheorem{assumption}{Assumption}

\def\paperID{6936} 
\def\confName{CVPR}
\def\confYear{2026}

\title{TWEO: Transformers Without Extreme Outliers Enables FP8 Training And Quantization For Dummies}

\author{
    Guang Liang$^{1,2,3}$ \quad \quad Jie Shao$^{1,2}$ \quad \quad Ningyuan Tang$^{1,2}$  \quad \quad 
    Xinyao Liu$^{4}$ \quad \quad Jianxin Wu$^{1,2}$\thanks{Corresponding author.} \\
    $^1$Nationa Key Laboratory for Novel Software Technology, Nanjing University, China \\
    $^2$School of Artificial Intelligence, Nanjing University, China \\
    $^3$Zhongguancun Academy, Beijing, China \\
    $^4$University of Science and Technology of China, Hefei, China \\
    {\tt\small{\{liangg, shaoj, tangny\}@lamda.nju.edu.cn, liuxinyao@mail.ustc.edu.cn, wujx2001@nju.edu.cn}}
}
\begin{document}
\maketitle

\begin{abstract}
Native FP8 support in modern hardware is essential for training large Transformers, but is severely hindered by extreme activation outliers. Existing solutions either rely on complex mixed-precision engineering or invasive architectural modifications. This paper fundamentally challenges the conventional wisdom that outliers are data-driven. We demonstrate that extreme outliers are a \emph{data-independent, mechanically-produced artifact of training}, originating from specific structural properties of the weight matrices (i.e., colinearity). Based on this insight, we propose TWEO (Transformers Without Extreme Outliers), a novel, non-invasive loss function. TWEO effectively prevents extreme outliers via a very simple loss term, which reduces outliers from 10000+ to less than 20. TWEO then enables \textbf{full-model FP8 pre-training} with neither engineering tricks nor architectural changes for both LLM and ViT. When standard FP8 training catastrophically collapses, TWEO achieves performance comparable to the BF16 baseline while delivering a 36\% increase in training throughput. Also, TWEO enables a new quantization paradigm. Hardware-friendly \textbf{W8A8 per-tensor static quantization} of LLMs, previously considered completely unusable due to outliers, achieves SOTA performance for the first time on TWEO-trained models.
\end{abstract}

\section{Introduction}
\label{sec:intro}

Computational demands of AI are growing at an unprecedented rate. In response, modern chips have introduced native support for 8-bit floating-point (FP8) computation, supported by software libraries such as the Transformer Engine~\cite{hernandez2025towards}, which aims to double training throughput while significantly reducing memory bandwidth requirements. But, it is still restricted by a core algorithmic barrier: the emergence of activation outliers, which are activation values whose magnitudes far exceed the average level~\cite{darcet2023vision, an2025systematic}.

Outliers are often categorized as ``normal'' and ``massive outliers''~\cite{sun2024massive, gallego2025hidden}. Normal outliers (e.g., around 10) may be unavoidable from multiplication of two high-dimensional vectors, which does not pose challenges for FP8 training or quantization. We focus on massive or extreme outliers (e.g., $>$1000)~\cite{dettmers2022gpt3}, which poses serious challenges.

First, in Post-Training Quantization (PTQ), outliers are the primary obstacle~\cite{bondarenko2023quantizable,yao2024exploring}. Extreme outliers drastically stretch the quantization range, forcing algorithms into a poor trade-off between clipping and rounding errors, which leads to severe accuracy drop~\cite{park2025outlier}. Experiments showed that simply zeroing out a few (0.1\%) outliers causes a catastrophic 600-1000\% spike in validation perplexity~\cite{dettmers2022gpt3}.

Second, in native low-bit training, they lead to catastrophic stability collapse. The range of FP8 is extremely limited (e.g., E4M3 ranges within $\pm 448$)~\cite{hernandez2025towards}. When extreme outliers ($>$1000) appear during training, numerical overflows are unavoidable and the failure mode (training loss suddenly explodes) will appear~\cite{liu2024deepseek}. This instability forces tricks such as reverting to BF16 or even FP32 for many sensitive components~\cite{qiu2025low}, which complicates the model design and largely removes FP8's efficiency gain.

On why extreme outliers emerge, existing research mostly linked them to properties of the input data, such as token frequency~\cite{puccetti2022outliers}, special tokens~\cite{chen2024prefixquant}, or sticky tokens~\cite{chen2025sticking}. Another data-dependent perspective is the no-op hypothesis~\cite{bondarenko2023quantizable}, which posits that the attention module attempts to learn a ``no-op'', which pushes the corresponding logits towards negative infinity, and hence requires the preceding layer to output very high magnitudes.

In this paper, we first find the cause of extreme outliers, then design a method to eliminate them. Following~\cite{shao2025reasons}, we design a \emph{Contradiction Stethoscope} to show that the root cause of extreme outliers is \emph{not data-dependent}. Instead, it is mostly data-independent, stemming from the interaction of network architecture, optimization algorithms, and the training dynamics. In vision, visual tokens (image patches) do not have an equivalent to high-frequency special tokens in LLM, but extreme outliers also emerge in Vision Transformers (ViT)~\cite{he2024understanding, ma2024outlier}. Thus, we suggest that the root cause of extreme outliers is structural, while data properties may further amplify it.

When the cause is ``mechanical'', so should be the solution for extreme outliers. We propose TWEO (Transformers Without Extreme Outliers), a loss function that directly penalizes extreme activation magnitudes. TWEO has the following core advantages:
\begin{itemize}
    \item \textbf{Universal}: It is data-independent, which directly suppresses extreme magnitudes regardless of their origin or functionality. Thus, TWEO is naturally cross-modal.
    \item \textbf{Flexible}: It can be plugged in to any Transformer architecture or variants.
    \item \textbf{Effective}: It eliminates loss explosion and ensures smooth and effective FP8 training.
    \item \textbf{Simple}: It can be used \emph{out of the box} (i.e., ``FP8 training for dummies''), without resorting to engineering tricks. Another benefit is that TWEO-trained models can be effectively quantized by the simplest method (i.e., ``quantization for dummies'').
\end{itemize}

\section{Related Work}
\label{sec:related_work}

Low-bit computation for Transformer models (FP8 training or quantization) are seriously hindered by extreme outliers. We briefly outline the solutions in the literature.

\textbf{Stabilize FP8 Training}. When extreme outliers exceeds the FP8 range, catastrophic collapse happens and disables FP8 training~\cite{gallego2025hidden, dettmers2022gpt3,liu2024deepseek}. Current solutions include
\begin{enumerate}
    \item Mixed-precision engineering~\cite{liu2024deepseek,team2025kimi,li2025every}. These methods find and keep sensitive modules (e.g., embeddings, normalization, MoE gates) in BF16, but entail high engineering costs and sacrifices FP8's efficiency benefits.
    
    \item Data-dependent architectural modifications. They introduce register tokens like ViT-R~\cite{darcet2023vision,chen2024prefixquant, puccetti2022outliers} to absorb outliers. But, we will show that extreme outliers are in fact data-independent.
    
    \item Data-independent invasive modifications. These methods propose invasive architectural changes such as modifying activation functions (Smooth-SwiGLU~\cite{fishman2024scaling}), attention(Clipped Softmax~\cite{bondarenko2023quantizable}), or completely redesigning the architecture (FOG~\cite{hernandez2025towards}).
\end{enumerate}
These solutions are either costly, based on inappropriate assumptions, or lack generality. The community needs a non-invasive, plug-and-play, universal solution for stable FP8 training without heavy engineering. We will propose TWEO to approaching this goal. 

\textbf{Simple and effective PTQ}. In Post-Training Quantization (PTQ,~\cite{cai2020zeroq,frantar2022gptq}), extreme outliers drastically stretch the quantization range and lead to poor accuracy~\cite{park2025outlier, yao2024exploring}. To address this challenge, existing methods (e.g., SmoothQuant, AWQ) have to transfer the difficulty. They concede that activations (especially the residual stream~\cite{saxena2024resq}) are difficult to quantize due to outliers, thus use mathematical transformations to transfer the quantization difficulty from the hard activations to the relatively easy weights. SmoothQuant~\cite{xiao2023smoothquant} introduces offline scaling factors, while AWQ~\cite{lin2024awq} scales activations by protecting salient weights. SpinQuant~\cite{liu2024spinquant} introduces learned rotation matrices to reduce the impact of outliers, while compensation methods attempt to fix or compensate for errors post-quantization~\cite{fu2025quantization}.

Although these methods have been effective to some extent in practice, they increase algorithmic and inference complexity. More importantly they are \emph{forced to keep the residual stream in high precision} (cf. Sec.~\ref{sec:exp_quant} and Appendix D), which introduces massive type-conversion operations and memory overhead and hinders acceleration in real-world scenarios.

These drawbacks stem from the fact they assume extreme outliers are unavoidable. Our TWEO eliminates extreme outliers, hence makes the simplest per-tensor static quantization viable (\emph{including the residual stream}), and renders complex methods (like SmoothQuant) unnecessary.

\section{Method}
\label{sec:method}

We start by examining existing assumptions on cause of extreme outliers, presenting our assumption, and then presenting the TWEO solution.

\subsection{Data-dependent Assumptions Are Problematic}

Existing work attribute extreme outliers to properties of the data, e.g., token frequency or special token~\cite{puccetti2022outliers,chen2024prefixquant,chen2025sticking,bondarenko2023quantizable}. Following~\cite{shao2025reasons}, we propose a \emph{Contradiction Stethoscope} to show that the root cause is \emph{not} data-dependent.

\begin{assumption}[The Contradiction Stethoscope]
Suppose the input data are changed to \emph{random numbers} (which do \emph{not} possess any of the special properties in data-dependent assumptions) but extreme outliers still persist at around the same scale, this contradiction will exclude data-dependency as the root cause.
\end{assumption}

\begin{figure*}[t]
    \centering
	\begin{minipage}{0.32\linewidth}
        \centering
        \includegraphics[width=\linewidth]{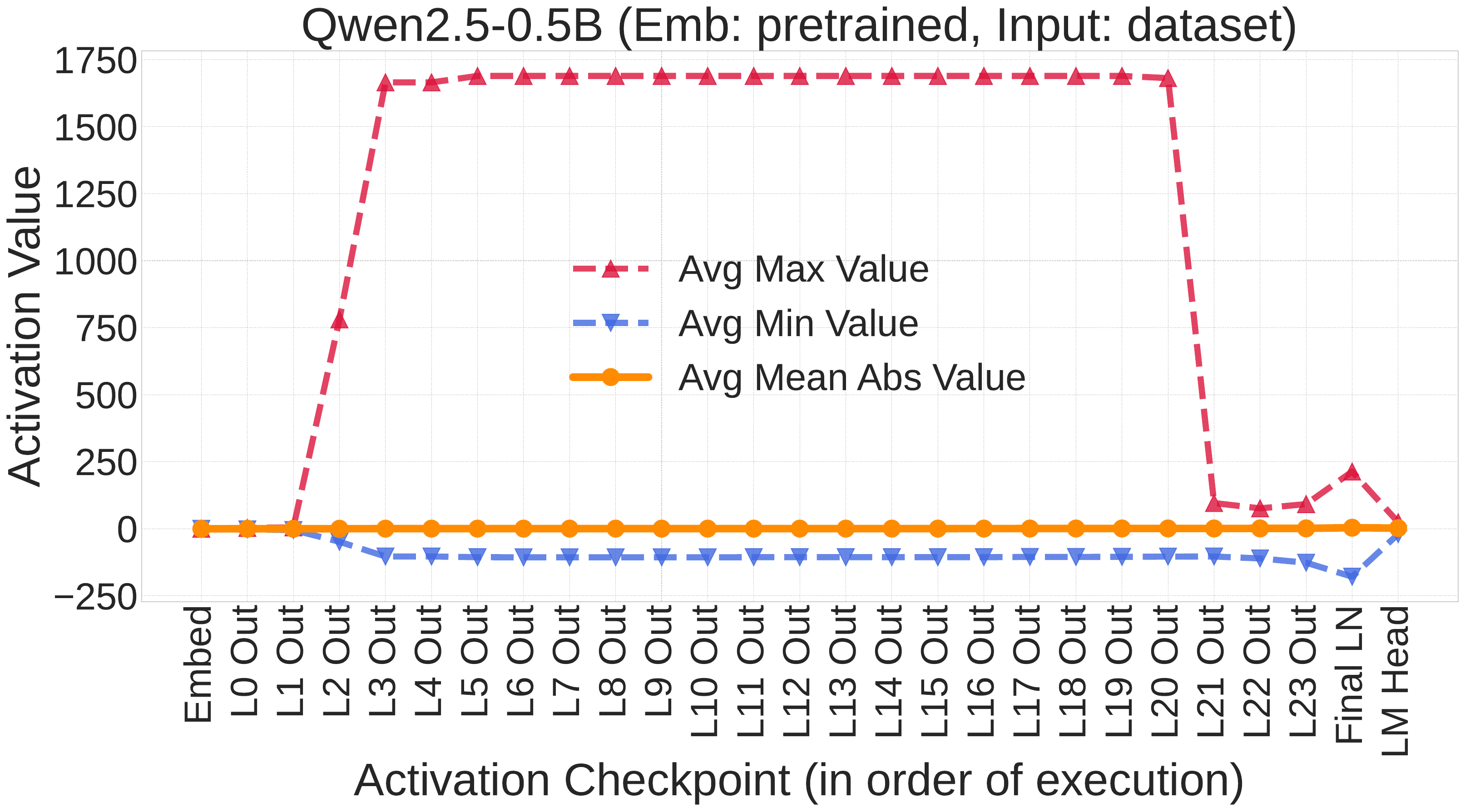}
        \small (a) Pre-trained model + Real data
    \end{minipage}
    \hfill 
	\begin{minipage}{0.32\linewidth}
        \centering
        \includegraphics[width=\linewidth]{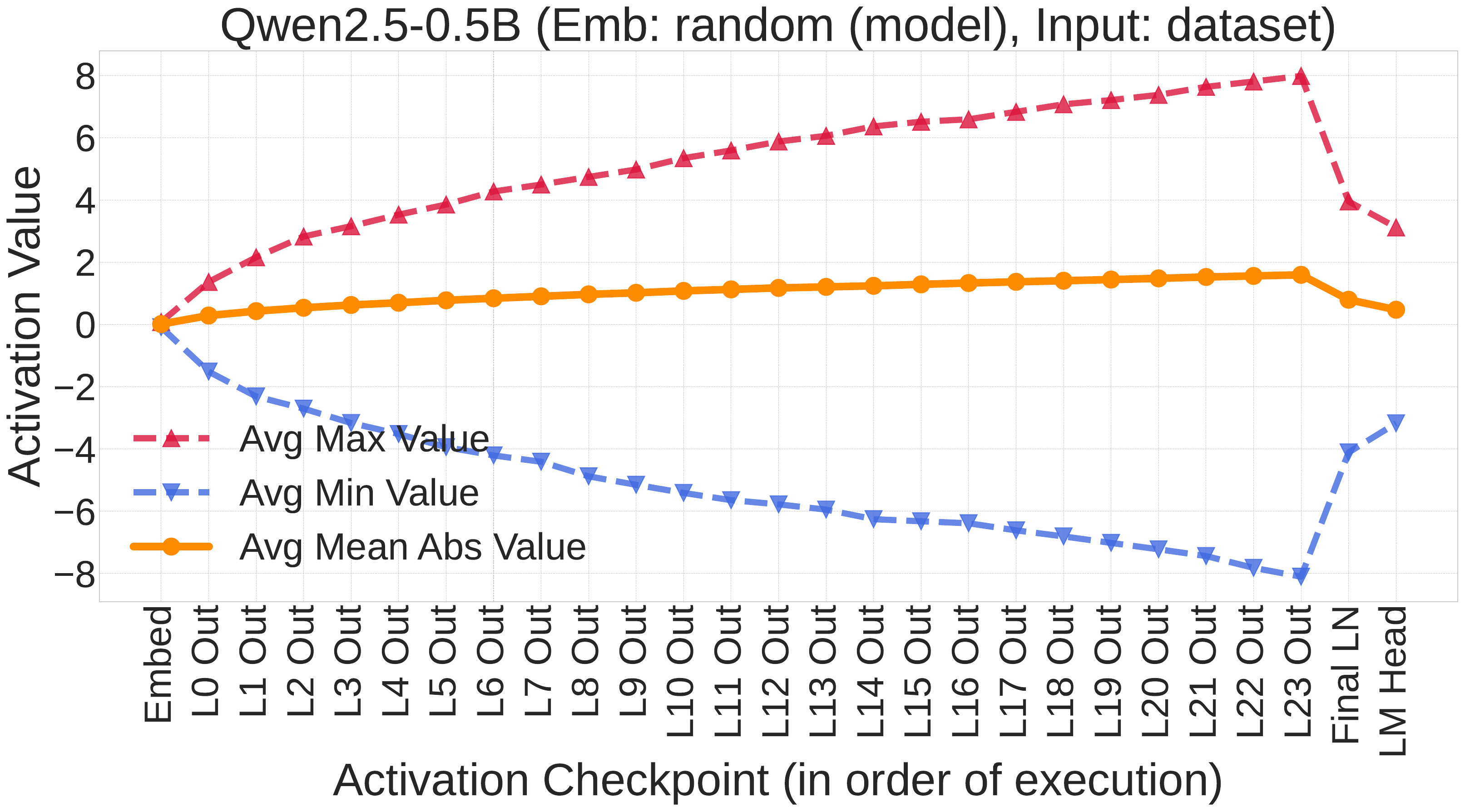}
        \small (b) Randomly initialized model + Real data
    \end{minipage}
    \hfill 
    \begin{minipage}{0.32\linewidth}
        \centering
        \includegraphics[width=\linewidth]{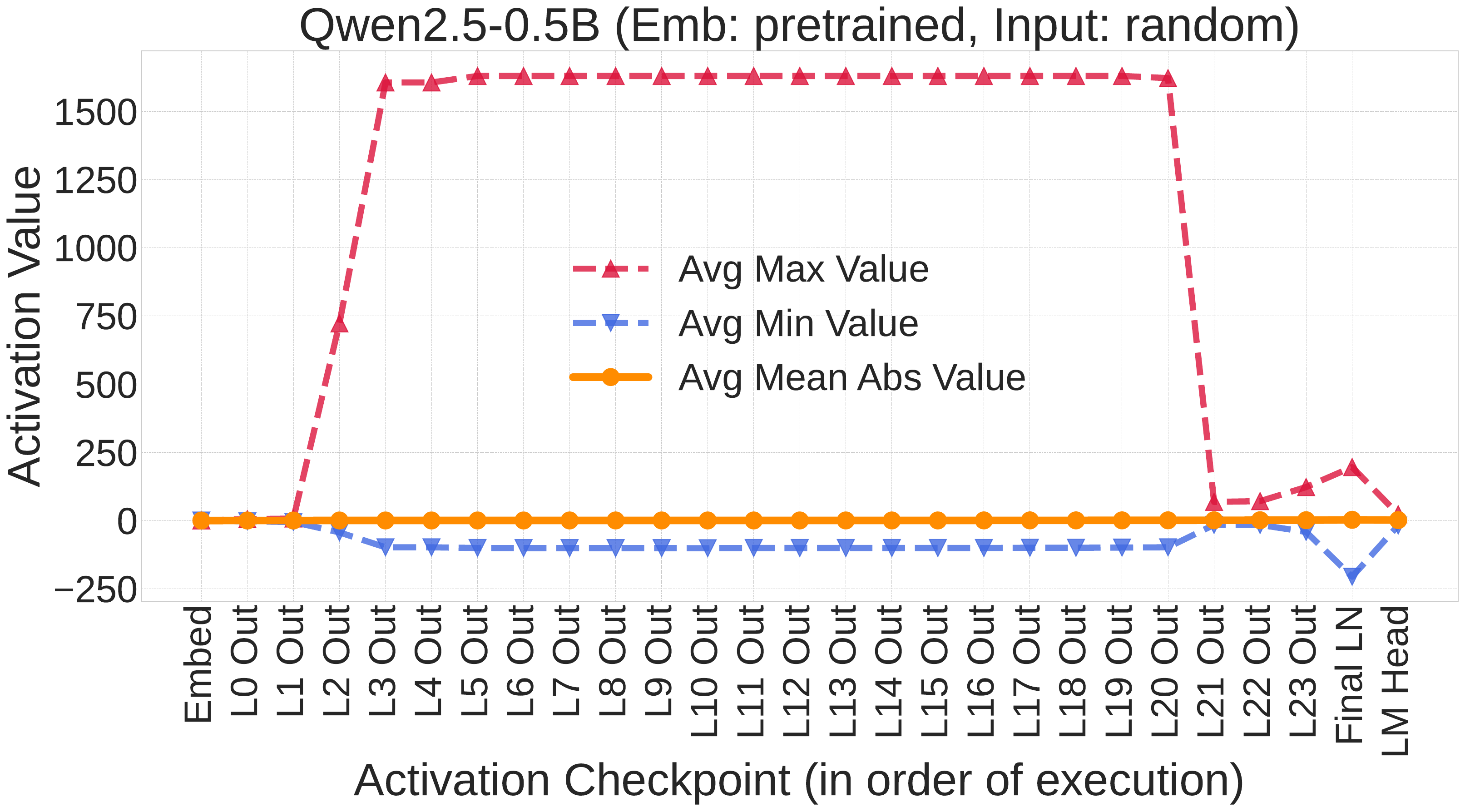}
        \small (c) Pre-trained model + Random data
    \end{minipage}

    \caption{Empirical evidences for data-dependent vs. data-independent assumptions. (a) Running real data through a pre-trained Qwen-0.5B LLM produces extreme outliers ($>$1000) in most layers. (b) But, real data through a \emph{randomly initialized} Qwen2.5-0.5B yields minimal activation magnitudes ($<$10). (c) And, changing input to \emph{random Gaussian noise}, the same pre-trained Qwen2.5-0.5B \textbf{still produces} extreme outliers.}
    \label{fig:data_independence_proof}
\end{figure*}

As shown in Figure~\ref{fig:data_independence_proof}, even when the input were changed to random numbers (which do not have special properties as in the data-dependent assumptions), the pre-trained model still produced extreme outliers. But, when real input data were sent to a \emph{randomly initialized} model, there was no extreme outliers at all. Hence, we assume that the root cause are \emph{not data-dependent}, but are associated with the \emph{trained weights}.

Hence, our data-independent assumption is
\begin{assumption}[Data-independent Root Cause]
Extreme outliers are not caused by properties of the input data, but are a mechanical artifact resulting from the weights' specific structures obtained during the training dynamics. 
\label{assumption:indep}
\end{assumption}

We provide more exhaustive studies span both language and vision in Appendix A, which examine the effects of model weights, token embeddings, and input data to extreme outliers.

\subsection{Colinearity Causes Extreme Outliers}
\label{sec:method_origin}

Recent studies~\cite{fishman2024scaling, he2024understanding,ma2024outlier} indicate that extreme outliers primarily originate in the MLP (Multi-Layer Perceptron) layers of the Transformer blocks. Hence, we conducted an empirical analysis of both classic MLPs with GELU activation\cite{hendrycks2016gaussian} and modern MLPs with GLU variants.

Consider a simplified classic MLP layer, whose computation can be approximately represented as $y=BAx$ by temporarily ignoring the activation function. We analyze the conditions under which the $k$-th element $y_k$ of the output $y \in \mathbb{R}^{d_1}$ becomes an outlier. $y_k$ can be expressed as the product of the $k$-th row of the matrix $B \in \mathbb{R}^{d_1 \times d_2}$ (denoted as $w^T \in \mathbb{R}^{1 \times d_2}$) and $Ax$.

Performing Singular Value Decomposition (SVD) on the matrix $A \in \mathbb{R}^{d_2 \times d_1}$ (up-projection) as $A=\sum_{i=1}^{d_{1}}s_{i}u_{i}v_{i}^{T}$, $y_k$ can be rewritten as
\begin{equation}
y_{k} = w^{T}Ax = \sum_{i=1}^{d_{1}}s_{i}(w^{T}u_{i})(v_{i}^{T}x) \,,
\label{eq:svd}
\end{equation}
which suggest a potential cause of outliers: when the row vector $w$ of matrix $B$ (down-projection) becomes \emph{collinear} with a left singular vector $u_i$ of matrix $A$, the product $(w^{T}u_{i})$ becomes very large. If, at the same time, the input $x$ is highly aligned with the corresponding right singular vector $v_i$ of $A$ (which causes $v_{i}^{T}x$ to be large), $y_k$ will become an extreme outlier due to the multiplication of these two large values.

We verified this hypothesis on a ViT-B model, ViT-B-21k which uses a classic GELU MLP. We located the first MLP layer to exhibit extreme outliers, with a value of 880. We then performed SVD analysis on the weight matrices $A$ and $B$ of this layer, i.e., $y = B \cdot \text{gelu}(A \cdot x)$. Analyzing only the dominant singular vector ($i=1$), we found that $s_1 = 18.189$, $w^T u_1 = -9.1847$ (which showed high weight and singular vector alignment), and $v_1^T x = -5.6165$ (which showed high input and singular vector alignment). Based on Eq.~\ref{eq:svd} (our analysis which ignores GELU), the simulated value is 938.28. When we further incorporated the GELU activation function, the simulated value became 884.29.

Note that the simulated value (884.29) is almost identical to the true outlier (880, relative error $<$0.5\%). Beyond this simple example, we find that whenever extreme outliers happened, Eq.~\ref{eq:svd} almost always simulated them accurately. This fact strongly suggests that outliers are a mechanical artifact resulting from structural alignment between the weight matrices ($A$, $B$) and the input ($x$).\footnote{When Eq.~\ref{eq:svd} predicts an extreme outlier, the real value may not be an extreme outlier because the activation function is ignored in Eq.~\ref{eq:svd}. That is, Eq.~\ref{eq:svd} predicts extreme outliers very accurately, but may produce false positives. False positives, however, are not important, because our Assumption~\ref{assumption:indep} focuses only on the root cause of extreme outliers.}

In architectures with the GLU activation function (e.g., in Llama\cite{touvron2023llama,touvron2023llama2,dubey2024llama}, Qwen\cite{bai2023qwen,yang2025qwen3}), the effect of Eq.~\ref{eq:svd} is further amplified by the (anti-)colinearity between the corresponding row vectors of the $\text{gate\_proj}$ and $\text{up\_proj}$ weight matrices~\cite{fishman2024scaling}, which leads to even large extreme outlier values. 

Combining the empirical evidences for Assumption~\ref{assumption:indep} and the theoretical analysis in Eq.~\ref{eq:svd}, our main hypothesis is: \textbf{extreme outliers are a structural artifact}. That is, they are not (or not solely) a direct representation of input data semantics (like special tokens), but are the result of mechanical amplification caused by specific properties that the network's weight matrices gradually develop during training dynamics (e.g., singular vector alignment or colinearity).

Then, we argue that to remove extreme outliers, we should not rely on modifying specific architectures (like the MLP or Attention), but design a method that directly suppresses this amplification effect during training mechanically.

\subsection{The Proposed TWEO Loss}
\label{sec:method_loss}

We then propose the TWEO (Transformers Without Extreme Outliers) loss $\mathcal{L}_{\text{TWEO}}$. It is a regularization term to the main task loss, $\mathcal{L}_{\text{task}}$:
\begin{equation}
\mathcal{L}_{\text{total}} = \mathcal{L}_{\text{task}} + \lambda(t) \mathcal{L}_{\text{TWEO}}\,,
\end{equation}
where the hyperparameter $\lambda(t)$ varies with training step $t$.

Unlike previous studies that might focus on internal MLP activations, our method intervenes at a more macroscopic scale. We monitor the final output activation $A^{(l)}$ of a Transformer block in different all blocks $l \in \{1, ..., L\}$, i.e., the $y$ after $y = x + \text{MLP}(\text{LN}(x))$.

$\mathcal{L}_{\text{TWEO}}$ is defined as the average scaled $L_p$ loss computed over the activations $A^{(l)}$ from all $L$ Transformer blocks:
\begin{equation}
\mathcal{L}_{\text{TWEO}} = \frac{1}{L} \sum_{l=1}^{L} \mathbb{E}\left[ \left( \frac{|A^{(l)}|}{\tau + \epsilon} \right)^p \right] \,,
\end{equation}
where $|\cdot|$ computes the absolute value, $A^{(l)}$ is the activation tensor of the $l$-th Transformer block, $\mathbb{E}[\cdot]$ denotes the mean (average) taken over the batch, sequence length, and hidden dimensions, $\tau > 0$ is a magnitude scaling factor or soft threshold, $p$ is the power of the penalty, and $\epsilon$ is a small constant (e.g., 1e-6) to ensure numerical stability.

The core of this design is to create a non-linear regularization mechanism that perfectly matches our goal of \emph{preserving normal values while suppressing extreme values}. This is achieved through the synergy of two key parameters: the magnitude scaling factor $\tau$ and the penalty power $p$ (where $\tau\in[2,5]$, $p=4$ in our design).

While the Hinge loss $\max(0, |A| - \tau)$ which uses a hard threshold, our $\tau$ is a soft threshold or a scaling factor. It does \emph{not} lead to discontinuous ``gradients'' so aids training stability, and it defines the target (desired) scale for activation magnitudes. 

The penalty power $p=4$ is key to achieving this non-linear penalty. The interaction of $\tau$ and $p$ is as follows.
\begin{itemize}
\item Tolerance for normal Values (when $|A| < \tau$). For activations much smaller than $\tau$ (e.g., $|A| = 0.5\tau$), the power $p=4$ renders its penalty term very small ($(0.5)^4 = 0.0625$). This allows the model to freely retain low-magnitude normal activations.

\item Setting the target scale (when $|A| = \tau$). When the activation equals the desired scale, the penalty for this term is a medium one: $(1)^p = 1$.

\item Suppression of extreme outliers (when $|A| > \tau$). For activations exceeding $\tau$ (e.g., $|A| = 10\tau\in[20,50]$), the penalty term is $(10)^4 = 10000$, which is a massive cost that will aggressively suppress extreme outliers.
\end{itemize}

TWEO is then highly effective at suppressing the extreme tail of the activation value's distribution, which is the root cause of FP8 overflows, while not affecting the training of normal activation values.

TWEO is simple and flexible, which allows it to be applied to various Transformer architectures. Because it acts directly on the physical magnitude of activations, rather than relying on specific task semantics (like token semantics in language or image structures in vision), it can be seamlessly extended to the training of vision, language, and even generative models (as shown in Appendix F).

\section{Experiments}
\label{sec:experiments}

We designed a series of experiments aimed at answering the following key questions: (1) Does TWEO possess cross-modal (vision, language) universality? (2) Can TWEO solve the loss explosion and instability issues in native FP8 pre-training? (3) What positive downstream impacts (e.g., on quantization) does TWEO have?

\subsection{Experimental Setup}
\paragraph{Models and Datasets}
Our experiments cover two mainstream modalities:
\begin{itemize}
    \item \textbf{Vision:} We trained Swin Transformers (Swin-T, Swin-S) \cite{liu2021swin} on the ImageNet-1K \cite{deng2009imagenet} dataset.
    \item \textbf{Language:} We pre-trained a series of GPT-2 models \cite{radford2019language} from scratch on the OpenWebText \cite{pile} dataset, with parameter scales ranging from 124M (GPT2) to 1.5B (GPT2-XL), as well as scaling to 3B (GPT2-3B) and 7B (GPT2-7B) using configurations similar to GPT-3 \cite{brown2020language}. Configuration details are available in the Appendix C.
\end{itemize}

We compared against BF16 training, standard native FP8 training (which led to training collapse), and outlier suppression methods in the literature~\cite{darcet2023vision, bondarenko2023quantizable}.

For TWEO, we applied it to the final output activations of all Transformer blocks. We found the settings for $\tau$ (scaling factor) and $p$ (penalty power) are robust across all tasks, with $\tau=3$ and $p=4$ in \emph{all our experiments} if not otherwise specified. $\lambda(t)$ was set to 0.01, optionally with a cosine annealing strategy.

\subsection{Results on Vision Tasks}
\label{sec:exp_vision}

\begin{table}
\centering
\caption{Results on Swin and ViT models on ImageNet. \#Params: parameter counts. Mod.: whether the network architecture has been modified. Peak: largest outlier value during training. Final: largest outlier value after training was finished.}
\label{tab:vision}
\setlength{\tabcolsep}{1pt}
\small
\begin{tabular}{lcccrr}
\toprule
\textbf{Method} & \textbf{Params} & \textbf{Mod.} & \textbf{Top-1 (\%)} & \textbf{Peak} & \textbf{Final} \\
\midrule
Swin-T~\cite{liu2021swin} (Baseline) & 28M & - & 81.2 & 1556 & 534 \\
Swin-T attn bias~\cite{an2025systematic} & 28M & Yes & 81.1 & 478 & 135 \\
Swin-T softmax+1~\cite{bondarenko2023quantizable,kaul2024attention} & 28M & Yes & 81.4 & 811 & 143 \\
Swin-T  gated~\cite{qiu2025gated} & 28M & Yes & 80.7 & 976 & 126 \\
Swin-T TWEO & 28M & No & \textbf{81.4} & \textbf{22} & \textbf{15} \\
\midrule
Swin-S~\cite{liu2021swin} (Baseline) & 50M & - & 82.7 & 6402 & 1758 \\
Swin-S TWEO & 50M & No & \textbf{82.8} & \textbf{22} & \textbf{10} \\
\midrule
Swin-B~\cite{liu2021swin} (Baseline) & 88M & - & 83.5 & 4824 & 876 \\
Swin-B TWEO & 88M & No & \textbf{83.4} & \textbf{23} & \textbf{8} \\
\midrule

ViT-S~\cite{dosovitskiy2020image} (Baseline) & 22M & - & \textbf{79.8} & 328 & 174 \\
ViT-S TWEO & 22M & No & 79.6 & \textbf{36} & \textbf{19} \\
\midrule
ViT-B~\cite{dosovitskiy2020image} (Baseline) & 87M & - & \textbf{81.3} & 1579 & 106 \\
ViT-B TWEO & 87M & No & \textbf{81.3} & \textbf{38} & \textbf{16} \\
\bottomrule
\end{tabular}
\end{table} 

\begin{figure*}
    \centering
    \begin{minipage}[t]{0.48\textwidth} 
        \centering
        \includegraphics[width=0.8\linewidth]{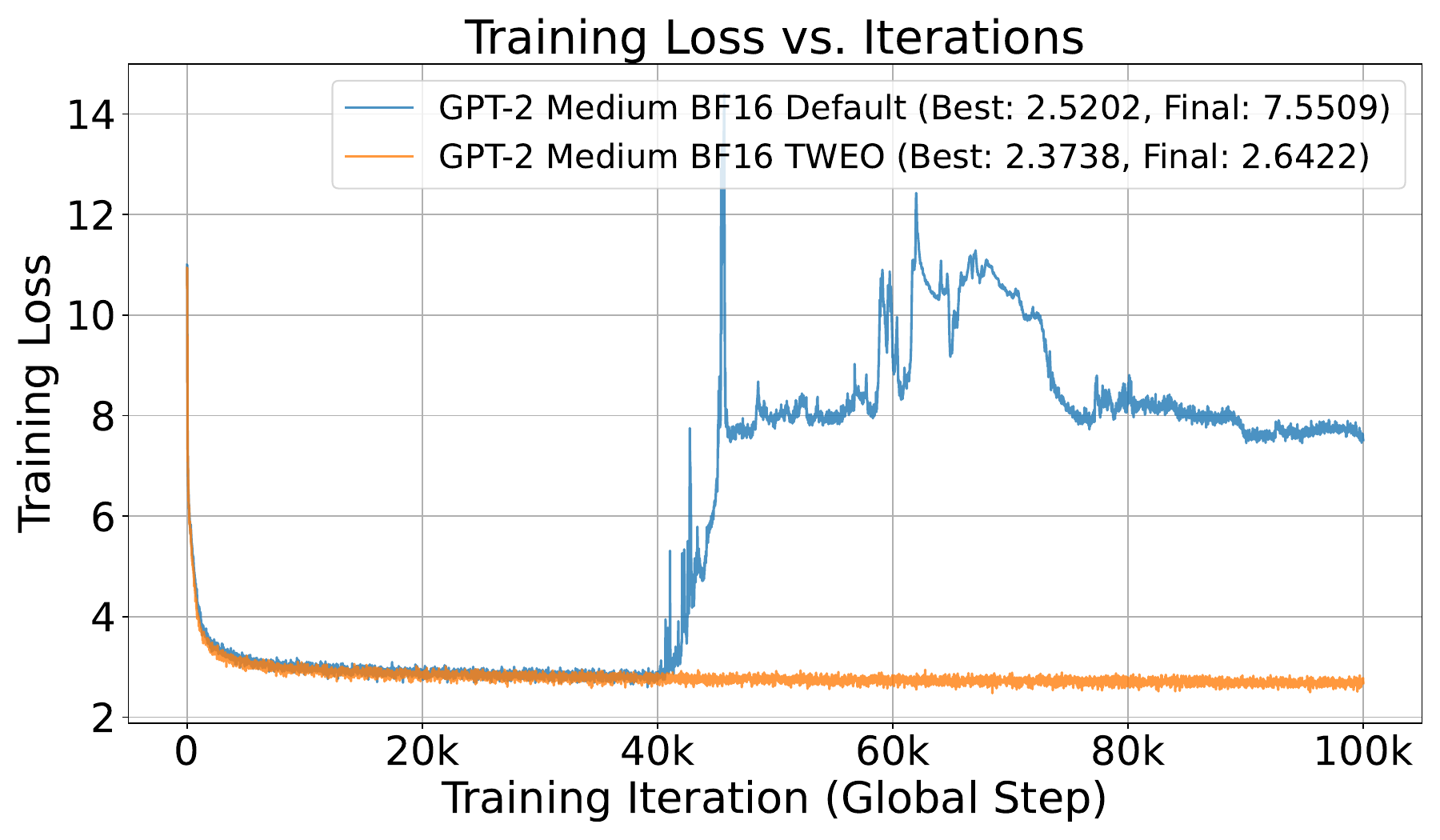} 
        \caption{Activation magnitudes in GPT-2 Medium BF16 training.}
\label{fig:training_loss_comparison_medium_bf16}
    \end{minipage}
    \hfill
    \begin{minipage}[t]{0.48\textwidth} 
        \centering
        \includegraphics[width=0.8\linewidth]{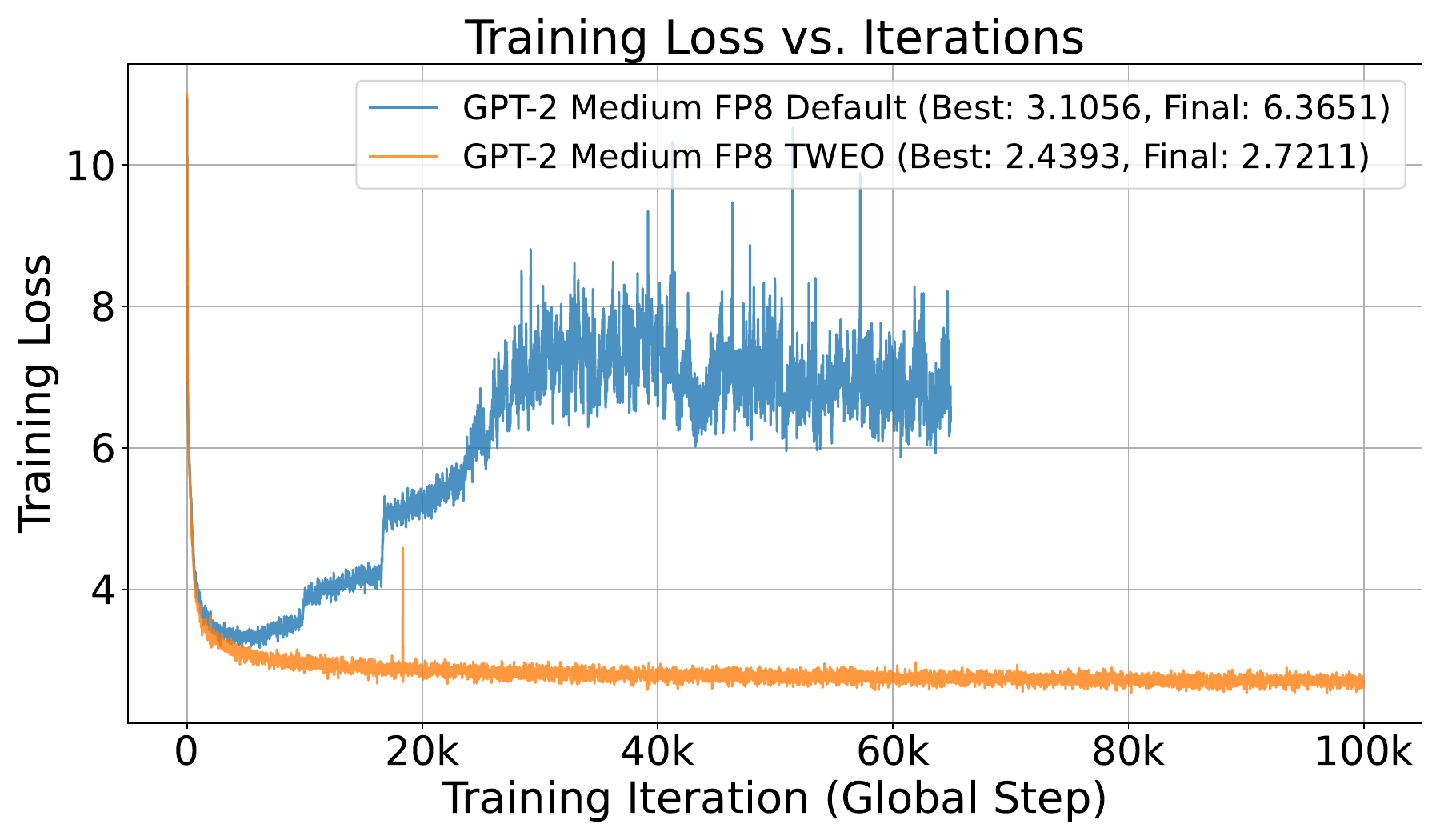} 
        \caption{Activation magnitudes in GPT-2 Medium FP8 training.}
\label{fig:training_loss_comparison_medium_fp8}
    \end{minipage}
\end{figure*}

We trained Swin Transformers and ViT on ImageNet and compared against invasive methods that require architectural changes. As shown in Table \ref{tab:vision}, we can observe
\begin{itemize}
    \item Existing methods require modifying the model's attention or activation modules. TWEO achieves similar accuracy to the baselines \emph{without any structural modifications}.
    \item TWEO eliminates extreme outliers while existing methods cannot. For Swin-S, it reduced the training peak magnitude from 6402 to 35.2 (99.4\% reduction), while for Swin-T from 1556 to 21.6 (98.6\% reduction).
\end{itemize}
This set of experiments confirm that the extreme outlier issue also exists in the vision models and that TWEO, as a data-independent method, can effectively solve it.

\subsection{LLMs: Stabilizing Full FP8 Pre-training}
\label{sec:exp_language}

This is the core application scenario for TWEO. To clearly demonstrate TWEO's simplicity and efficiency, we compare it with the FP8 training solution of the well-known DeepSeek-V3~\cite{liu2024deepseek} model.

\begin{table}
\centering
\small
\caption{FP8 pre-training. PPL: Validation perplexity. $\downarrow$: lower is better. Peak: largest outlier during training.}

\label{tab:gpt2_pretrain}
\begin{tabular}{@{}llrrr@{}}
\toprule[0.45mm]
\textbf{Model} & \textbf{Params} & \begin{tabular}[c]{@{}c@{}}\textbf{PPL}\\ \textbf{(FP8)}\end{tabular} & \begin{tabular}[c]{@{}c@{}}\textbf{PPL}\\ \textbf{(BF16)}\end{tabular} & \begin{tabular}[c]{@{}c@{}} \textbf{Peak}\end{tabular} \\ 
\midrule[0.25mm]
GPT2 (baseline) & 124M & 169.81 & 20.04 & 823 \\

Attn bias \textsuperscript{*} \cite{an2025systematic} & 124M & 24.93 & 19.79 & 163 \\

Attn ca-softmax \textsuperscript{*} \cite{an2025systematic} & 124M & 23.81 & 19.34 & 135 \\

Baseline+TWEO & 124M & \textbf{19.26} & \textbf{18.68} & \textbf{17} \\ 
\toprule[0.25mm]
GPT2-medium & 350M & 127.34 & 16.77 \textsuperscript{\#} & 24563 \\

\quad +TWEO & 350M & \textbf{15.64} & \textbf{15.18} & \textbf{19} \\ 
\toprule[0.25mm]
GPT2-large & 774M & 178.66 & 14.78 \textsuperscript{\#} & 32361 \\
\quad +TWEO & 774M & \textbf{13.89} & \textbf{13.79} & \textbf{18} \\ 
\toprule[0.25mm]
GPT2-xl & 1.6B & 93.28 & 13.84 & 32889 \\
\quad +TWEO & 1.6B & \textbf{12.58} & \textbf{12.39} & \textbf{19} \\ 
\toprule[0.25mm]
GPT2 3B & 3B & 76.85 & - & - \\

\quad +TWEO & 3B & \textbf{12.24} & - & \textbf{18} \\ \toprule[0.25mm]
GPT2 7B & 7B & 67.12 & - & - \\

\quad +TWEO & 7B & \textbf{12.02} & - & \textbf{20} \\ \bottomrule[0.45mm]
\end{tabular}%
\vspace{-1.0mm}
\begin{flushleft} 
\footnotesize
\textsuperscript{*} Invasive methods.

Attn bias: Adds learnable biases $k'$ and $v'$ to the Key and Value matrices.

Attn ca-softmax: Adds a learnable, context-dependent scaling factor.

\textsuperscript{\#} Training collapsed. We report the best checkpoint before collapsing.
\end{flushleft}
\vspace{-3.8mm}

\end{table}

DeepSeek-V3 stated that low-precision training is ``limited by the presence of outliers in activations, weights, and gradients''~\cite{liu2024deepseek}, hence it is is a highly complex, deeply customized \emph{mixed-precision} engineering effort:
\begin{itemize}
    \item Bypassing Outliers: It is forced to keep all outlier-prone areas, such as the embedding module, language model head, and normalization operators in BF16 or FP32.
    \item Isolating Outliers: For modules using FP8, it relies on \textbf{fine-grained} `tile-wise' or `block-wise' quantization strategies, calculating separate scaling factors for different blocks of a tensor to isolate local outliers.
\end{itemize}
This approach requires many customized operators (e.g., online quantization and CUDA Core accumulation~\cite{liu2024deepseek}), making it difficult to transfer and popularize.

\begin{figure*}
    \centering
    \includegraphics[width=0.8\textwidth]{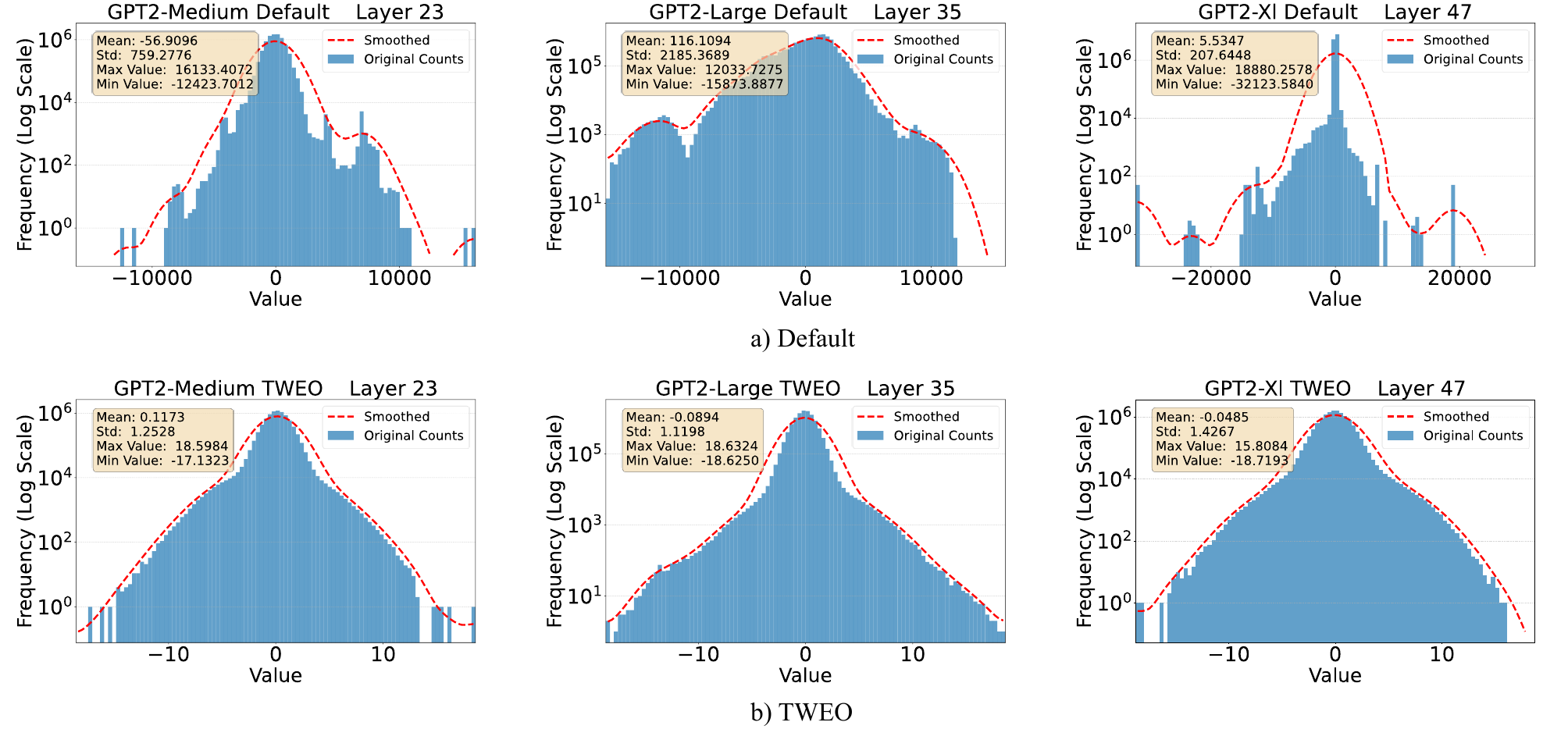}
    \caption{Activation value distributions for various GPT-2 model sizes, with and without TWEO.}
    \label{fig:gpt2_layer_outliner_hist}
\end{figure*}

\begin{figure}
    \centering
  \includegraphics[width=\columnwidth]{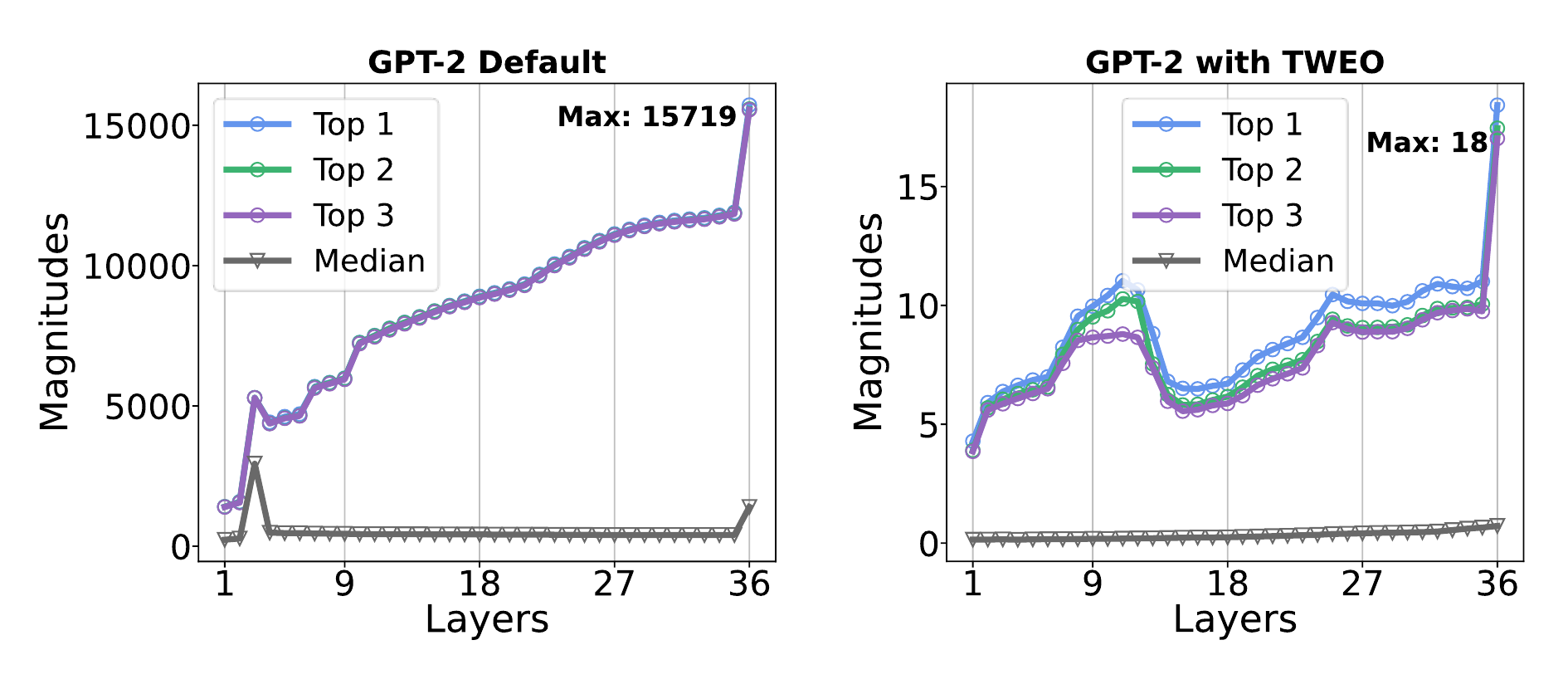}
    \caption{Comparison of activation magnitudes during GPT-2 Large (774M) training. Top 1 refers to the largest activation value in each layer. Top 2, Top 3 refer to the second and third largest. Median is the median of all activations.}
    \label{fig:gpt2_774m}
\end{figure}

In contrast, the core advantage of our TWEO is that it \emph{eliminates outliers from the root cause} (cf. Figures~\ref{fig:gpt2_774m} and~\ref{fig:gpt2_layer_outliner_hist}). This obviates the need for complex engineering workarounds to adapt to or bypass outliers. Hence, we are able to adopt an \emph{very aggressive} and \emph{minimalist} training strategy. We placed all Linear layers and LayerNorm components (including the final language model head) under the FP8 automatic mixed precision (autocast) context for computation—--a setup that is unacceptable in frameworks like DeepSeek-V3.

Our FP8 training setup aims for maximum simplicity. We utilize the NVIDIA Transformer Engine~\cite{hernandez2025towards}, employing the \emph{coarse-grained} \texttt{DelayedScaling} strategy, which has the lowest hardware overhead but \textbf{zero tolerance} for outliers. It uses \emph{one single scaling factor for the entire tensor} (true per-tensor scaling) and the \texttt{Format.HYBRID} (E4M3/E5M2) format.

Its aggressive nature is particularly evident in our configuration of an \emph{extremely short} \texttt{amax\_history\_len=16}. This is a high-risk configuration: it assumes a stable activation distribution (since `per-tensor' scaling cannot isolate local spikes) and is thus extremely prone to training collapse when an outlier emerges.

As Figures~\ref{fig:training_loss_comparison_medium_bf16} and~\ref{fig:training_loss_comparison_medium_fp8} show, this high-risk setup perfectly validates our hypothesis. The standard model (Baseline) suffered a catastrophic loss explosion under this stringent configuration. In contrast, TWEO kept a smooth and stable convergence trajectory identical to the BF16 baseline, even under this simplest, most aggressive per-tensor FP8 setting.

Table~\ref{tab:gpt2_pretrain} summarizes these experiments. Standard FP8 training (Baseline) suffered catastrophic collapse across all model scales from 124M to 7B. Its validation perplexity (PPL) is extremely high (e.g., 193.2 for GPT2-XL, 183.1 for GPT2 7B). In contrast, TWEO (`+TWEO' in Table~\ref{tab:gpt2_pretrain}) not only successfully completed training at all scales, but their FP8 PPL matched or even surpassed those of the BF16 baseline (e.g., 13.01 for GPT2-XL, 12.02 for GPT2 7B). The successful suppression of extreme outliers is the key: as shown in the last column, the baseline models' peak magnitudes reached tens of thousands, whereas TWEO consistently controlled the peak magnitudes of all models to \textbf{$\le$20}.

Figure~\ref{fig:gpt2_layer_outliner_hist} visualizes TWEO's outlier suppression effect from a distributional perspective. The baseline model (`Default') had many modes and is heavy-tailed, whose extreme values (e.g., $\pm 30000$ in GPT2-XL) caused the distribution to be extremely stretched. In contrast, TWEO's activation distributions of are single mode, with a mean near 0 and small standard deviation, and eliminated outliers.

Figure~\ref{fig:gpt2_774m} intuitively demonstrates TWEO's mechanical suppression effect. The baseline model's training peak magnitude reached 30000, while TWEO's activation values across \emph{all layers} were smoothly clamped below 20.

\subsection{Post Training: A New Quantization Landscape}
\label{sec:exp_quant}

\begin{table*}
\centering
\caption{Perplexity (PPL) comparison of Baseline vs. TWEO models under different 8-bit AbsMax PTQ strategies on OpenWebText.}
\label{tab:quant_absmax_ptq}
\footnotesize
\setlength{\tabcolsep}{1pt}
\begin{tabular}{llc | cc | cc}
\toprule
\textbf{Model (Params)} & \textbf{Method} & \textbf{BF16 PPL} & \begin{tabular}[c]{@{}c@{}}\textbf{Act. Only 8-bit (A8)} \\ per-tensor / per-token\end{tabular} & \begin{tabular}[c]{@{}c@{}}\textbf{Weight Only 8-bit (W8)} \\ per-tensor / per-channel\end{tabular} & \begin{tabular}[c]{@{}c@{}}\textbf{Full 8-bit Quant (W8A8)} \\ W8(T)A8(T) / W8(T)A8(K)\end{tabular} & \begin{tabular}[c]{@{}c@{}}\textbf{Full 8-bit Quant (W8A8)} \\ W8(C)A8(T) / W8(C)A8(K)\end{tabular} \\
\midrule
\multirow{2}{*}{GPT-2 (124M)} & Default& 20.04 & 82.94 / 21.19 & 20.71 / 20.32 & 86.60 / 22.02 & 82.94 / 21.49 \\
 & \textbf{TWEO)} & \textbf{18.83} & \textbf{20.21} / \textbf{19.00} & \textbf{19.32} / \textbf{19.11} & \textbf{20.82} / \textbf{19.51} & \textbf{20.54} / \textbf{19.00} \\
\midrule
\multirow{3}{*}{GPT-2 Medium (350M)} & Default& 16.77 & 1451.40 / 19.76 & 17.35 / 17.17 & 1491.11 / 20.24 & 1456.97 / 19.95 \\
 & \textbf{TWEO} & \textbf{15.18} & \textbf{16.28} / \textbf{15.34} & \textbf{15.38} / \textbf{15.25} & \textbf{16.50} / \textbf{15.53} & \textbf{16.32} / \textbf{15.40} \\
\midrule
\multirow{2}{*}{GPT-2 Large (0.8B)} & Default& 14.92 & 43204.39 / 23.20 & 2228.19 / 2228.19 & 43720.76 / 24.14 & 45821.84 / 23.35 \\
 & \textbf{TWEO} & \textbf{13.79} & \textbf{18.09} / \textbf{13.98} & \textbf{13.98} / \textbf{13.89} & \textbf{18.88} / \textbf{14.17} & \textbf{18.39} / \textbf{14.07} \\
\midrule
\multirow{2}{*}{GPT-2 XL (1.5B)} & Default& 13.84 & 1799.44 / 22.42 & 14.27 / 14.08 & 1872.83 / 22.01 & 1799.44 / 23.35 \\
 & \textbf{TWEO} & \textbf{12.39} & \textbf{12.86} / \textbf{12.43} & \textbf{12.61} / \textbf{12.58} & \textbf{13.09} / \textbf{12.66} & \textbf{13.06} / \textbf{12.63} \\
\bottomrule
\end{tabular}%
\end{table*}

The value of TWEO lies not only in enabling stable low-bit training but also in facilitating post-training quantization. A consensus is that extreme activation outliers are the major barrier to quantization~\cite{yao2024exploring, liang2025gplq}. This issue has led the field to \emph{abandon the use of per-tensor static quantization for activations}, because per-tensor quantization is the most hardware-friendly and fastest inference solution, but it is also the most sensitive to outliers. We have been forced to use per-token or other more expensive quantization schemes.

Now that TWEO has removed this barrier (extreme outliers), it must enable simple and effective post-training quantization. To verify this, we applied the strictest (most outlier-sensitive) and simplest (most hardware-friendly) \emph{AbsMax static quantization} to both baseline and TWEO-trained models. AbsMax static quantization is simple, symmetric, and hardware-friendly (zero-point-free). We use the "W$b$A$a$" notation (e.g., ``W8A8'') to denote weights quantized to $b$ bits and activations to $a$ bits.

For a $b$-bit symmetric quantization, we first define the maximum quantization integer $Q_b = 2^{b-1} - 1$. For a floating-point tensor $X$, the quantization process is:
\begin{equation}
s = \max(|X|) + \epsilon\,, X_q = \text{round}\left( \frac{X}{s} \cdot Q_b \right) \,,
X_{\text{deq}} = \frac{X_q}{Q_b} \cdot s 
\end{equation}
The speed bottleneck lies in the granularity of the scaling factor $s$ calculation, as:
\begin{itemize}
    \item Per-tensor (T): (Used for W and A) A single scaling factor $s$ for the \emph{entire} tensor.    
    \item Per-channel (C): (Used for W only) Independent $s_c$ for \emph{each} output channel.
    \item Per-token (K): (Used for A only) Independent $s_t$ for \emph{each token vector}. Highest overhead.
\end{itemize}
We compare combinations of these granularities to comprehensively evaluate TWEO's impact on PTQ.

Results in Table~\ref{tab:quant_absmax_ptq} clearly demonstrate a paradigm shift:
\begin{itemize}
    \item As expected, the Default (Baseline) models completely collapsed in performance whenever per-tensor activation quantization, A(T) or W(T)+A(T), was involved. For example, the PPL of GPT-2 Medium spiked from 16.77 to 1451.40 or 1491.11.
    \item But, activations are no longer the bottleneck in TWEO. On GPT-2 XL trained by TWEO, the per-token A8(K) quantization PPL (12.43) is even better than its per-channel W8(C) PPL (12.58). This observation subverts the traditional wisdom ``activations are always harder to quantize than weights.''
    \item The critical finding comes from the W(T)+A(T) (\emph{full per-tensor}) strategy. For GPT-2 Medium, TWEO achieved a PPL of 16.50. This result is significant because a) It stands in stark contrast to the baseline model's PPL (1491.11), and b) It is even better than the Default (Baseline)'s BF16 PPL (16.77). This result show that by eliminating outliers with TWEO, we \emph{may not need} the complex difficulty transfer techniques of SmoothQuant or SpinQuant. Instead, we just need to directly employ the fastest, simplest-to-implement per-tensor static quantization scheme while achieving low performance loss.
\end{itemize}

Table~\ref{tab:quant_absmax_ptq} proves that TWEO can make simple AbsMax quantization viable, but a more profound problem lies in the quantization of the \textbf{residual stream}. In the Transformer architecture, activations are passed between layers via a residue stream $y = x + f(x)$. Extreme outliers are passed and accumulated in the residual stream layer by layer. Consequently, difficulty transfer methods like SmoothQuant are forced to operate \emph{only} inside $f(x)$ while being compelled to leave $x$ itself (the residual stream) in high precision (BF16/FP32). As shown in Table~\ref{tab:quant_residual_xl_main}, quantizing $x$ led to catastrophic collapse. This compromise (not quantizing the residual) results in frequent BF16 $\leftrightarrow$ int8 type conversions and introduces significant memory and latency overhead, severely hampering inference speed.

\begin{table}
\centering
\caption{Perplexity of GPT-2 XL. SmoothQuant collapsed when the residual was quantized. More details are in Appendix D. Res.?: whether the residual stream was quantized or not.}
\label{tab:quant_residual_xl_main}
\footnotesize
\setlength{\tabcolsep}{1.5pt}
\begin{tabular}{lccc}
\toprule
\textbf{Model PPL (BF16)} & \textbf{Method} & \textbf{Res.?} & \textbf{W8(C)A8(T) / W8(C)A8(K)} \\
\midrule

\multirow{2}{*}{\makecell[l]{GPT-2 XL Default \\ (PPL: 13.84)}} & \cellcolor{gray!25}SmoothQuant & \cellcolor{gray!25}No & \cellcolor{gray!25}14.81 / 14.01 \\
 & SmoothQuant & \textbf{Yes} & 1876.70 / 21.93 \\
\midrule
\multirow{2}{*}{\makecell[l]{GPT-2 XL TWEO \\ (PPL: 12.39)}} & \textbf{AbsMax} & \textbf{Yes} & 13.06 / 12.63 \\
 & \textbf{SmoothQuant} & \textbf{Yes} & \textbf{12.89} / \textbf{12.51} \\
\bottomrule
\end{tabular}%
\end{table}

By eliminating extreme outliers, TWEO makes the quantization of the residual stream $x$ possible \emph{for the first time}. Results in Table~\ref{tab:quant_residual_xl_main} show that
\begin{itemize}
    \item When SmoothQuant was forced to quantize the residual stream, its W8(C)A8(T) PPL on the Default model catastrophically collapsed from 14.81 to 1876.70.
    \item Conversely, the TWEO model, even when using AbsMax and quantizing the residual stream, maintained a PPL of 13.06 / 12.63, which is even better than all baselines.
    \item On the TWEO pre-trained model, applying the complicated SmoothQuant method yielded negligible gains compared to simple AbsMax. This shows that TWEO has eased difficulties in quantization of \emph{both} activations and weights, making complex difficulty transfer methods unnecessary.
\end{itemize}
That is, our work introduces a new paradigm for post-training quantization, which makes full-model quantization (\emph{including the residual stream}) possible and paves the way for unleashing the full potential of low-bit inference.

\textbf{Simple quantization of vision models}. As a general outlier suppression method, TWEO's positive impact on downstream quantization is also applicable in computer vision tasks. We conducted parallel PTQ experiments on vision Transformers (ViT and Swin), using per-channel static AbsMax quantization for weights and the most outlier-sensitive \emph{per-tensor} scheme for activations (W(C)A(T)), with results shown in Table~\ref{tab:vision_quant}.

\begin{table}
\centering
\small
\caption{TWEO's impact on Vision Transformer PTQ. All quantization uses \emph{static AbsMax quantization with per-tensor activations (W(C)A(T))}. The table clearly shows baseline models collapsed at W6A6, while TWEO successfully recovered accuracy to highly competitive levels.}
\label{tab:vision_quant}
\setlength{\tabcolsep}{1.5pt}
\begin{tabular}{llccc}
\toprule
\textbf{Model} & \textbf{Method} & \begin{tabular}[c]{@{}c@{}}\textbf{W32A32} \\ \textbf{ (\%)}\end{tabular} & \begin{tabular}[c]{@{}c@{}}\textbf{W8A8} \\ \textbf{(\%)}\end{tabular} & \begin{tabular}[c]{@{}c@{}}\textbf{W6A6} \\ \textbf{(\%)}\end{tabular} \\
\midrule
\multirow{2}{*}{ViT-B (87M)} & Default & 81.32 & 79.86 & 7.41 \\
 & \textbf{+TWEO (Ours)} & \textbf{81.22} & \textbf{80.29} & \textbf{66.37} \\
\midrule
\multirow{2}{*}{Swin-T (28M)} & Default & 81.24 & 77.69 & 0.19 \\
 & \textbf{+TWEO (Ours)} & \textbf{81.40} & \textbf{80.94} & \textbf{51.69} \\
\midrule
\multirow{2}{*}{Swin-S (50M)} & Default & 82.74 & 80.22 & 0.13 \\
 & \textbf{+TWEO (Ours)} & \textbf{82.79} & \textbf{82.55} & \textbf{77.27} \\
\midrule
\multirow{2}{*}{Swin-B (88M)} & Default & \textbf{83.53} & 40.16 & 0.10 \\
 & \textbf{+TWEO (Ours)} & 83.41 & \textbf{83.08} & \textbf{80.77} \\
\bottomrule
\end{tabular}%
\end{table}

Results in Table~\ref{tab:vision_quant} again validate TWEO's generality. In the W8A8 setting, TWEO-trained models significantly outperformed the baselines. In the more aggressive W6A6 setting, the baseline models completely collapsed (e.g., Swin-T at 0.19\%, ViT-B at 7.41\% accuracy), whereas TWEO-trained models successfully recovered the accuracy to highly competitive levels (Swin-S at 77.27\%, Swin-B at 80.77\%). This proves that TWEO, as the first general outlier solution effective in both vision and language, creates an outlier-free model paradigm that provides advantages for low-bit quantization across domains.

\textbf{Implications for the Quantization Field}. That is, TWEO has the potential open up a new, outlier-free research direction for the field of quantization.

For academia research, the future focus may shift from ``how to design complex algorithms to bypass outliers'' to ``with an outlier-free model, how low-bit can we go?'' TWEO simplifies the quantization problem itself, allowing researchers to focus on exploring more aggressive low-bit (e.g., W4A8, W4A4) compression schemes for this clean (i.e., outlier free) class of models.

For industry applications, a scheme using per-tensor static quantization for both activations and weights such as W8(T)A8(T) is unrivaled in its appeal. It requires no complex re-scaling or online rotations, supports maximum concurrency, and enjoys minimum latency, power consumption, and memory footprint. TWEO proves that this simplest, most efficient path is in fact viable, and the substantially reduced costs can drive on-device deployment of more artificial intelligence products.

Furthermore, TWEO opens new paths for hardware-software co-design. Previous ASICs (Application-Specific Integrated Circuits) had to build costly control logic for complex per-token/per-channel quantization schemes (to circumvent the outlier challenges~\cite{xiao2023smoothquant}). This not only increases chip area and power consumption but also restricts their maximum clock frequency~\cite{sze2017efficient,wang2019haq}.

TWEO's potential contribution in this aspect is making the simplest, most hardware-friendly \emph{static per-tensor quantization} viable again (as shown in Table~\ref{tab:quant_absmax_ptq}). This is a significant benefit for next-generation specialized AI accelerators: designers can remove complex quantization logic and instead design leaner, more specialized computation cores. This simplified design has the potential to translate into higher clock frequencies at lower power, enabling a leap in energy efficiency~\cite{sze2017efficient,tensorrt2023}.

\section{Conclusions}
\label{sec:conclusion}

In this paper, we confronted one of the core bottlenecks hindering next-generation AI hardware from reaching its full potential: extreme activation outliers. Instead of treating them as symptoms to be bypassed, we hypothesized that their root cause is \emph{data-independent, mechanical artifacts} of the training dynamics. Then we designed TWEO, a plug-and-play, non-invasive loss function, which cures the model by proactively suppressing the formation of extreme magnitudes at every step of training.

Our experiments on both language and vision up to 7B models demonstrated the impact of this approach:
\begin{itemize}
    \item \emph{In FP8 Training}: It transforms catastrophic, unreliable full FP8 training into a stable, efficient (+36\% throughput) process with performance on par with BF16.
    \item \emph{In Post-Training Quantization}: It allows the field to move beyond its dependency on complex difficulty transfer algorithms (like SmoothQuant). We demonstrated that on an outlier-free model produced by TWEO, the most hardware-friendly, simplest \emph{per-tensor static quantization}—including \emph{full residual stream} quantization—becomes viable and effective for the first time.
\end{itemize}
TWEO also moves low-bit computation from expert customization to out-of-the-box technology.

Due to resource constraints, we had not verified TWEO on the largest models (e.g., 700B), which will be a future work when compute is available. While TWEO models were trained from scratch in this paper, we are also interested in fine-tuning existing models to remove their extreme outliers. And, we are interested in further FP4 learning and inference.




\section*{Acknowledgments}
This work was supported by the National Natural Science Foundation of China under Grant 62276123. We also thank Zhongguancun Academy for providing the computing resources.

J.W. designed the strategy to eliminate extreme outliers via finding its root cause, and then enables easy FP8 training and quantization. J.W., with the help of G.L., proposed the \textit{Contradiction Stethoscope}. G.L., with the help of J.W., captured the data independent nature of outliers, proposed the TWEO loss, and successfully implemented it.  J.S., N.T., and X.L. provided essential support in carrying out the experiments and analyses. All authors contributed to the final manuscript.

{
    \small
    \bibliographystyle{ieeenat_fullname}
    \bibliography{main}
}



\end{document}